\documentclass{article}

\usepackage{multirow}
\usepackage{graphicx}
\usepackage{amsmath}
\usepackage{subcaption}
\usepackage{amssymb}% http://ctan.org/pkg/amssymb
\usepackage{pifont}% http://ctan.org/pkg/pifont
\newcommand{\cmark}{\ding{51}}%
\newcommand{\xmark}{\ding{55}}%
\usepackage[preprint]{neurips_2026}

\usepackage[utf8]{inputenc} % allow utf-8 input
\usepackage[T1]{fontenc}    % use 8-bit T1 fonts
\usepackage{hyperref}       % hyperlinks
\usepackage{url}            % simple URL typesetting
\usepackage{booktabs}       % professional-quality tables
\usepackage{amsfonts}       % blackboard math symbols
\usepackage{nicefrac}       % compact symbols for 1/2, etc.
\usepackage{microtype}      % microtypography
\usepackage{xcolor}         % colors
\usepackage{comment}

\newcommand{\todo}[1]{{\color{red} \textbf{[TODO: #1]}}}
\renewcommand{\todo}[1]{}

\title{How Should LLMs Listen While Speaking? A Study of User-Stream Routing in Full-Duplex Spoken Dialogue}

\author{%
  Hui Lu$^{1}$\thanks{Work done during an internship at SenseTime Research} , 
  Xueyuan Chen$^1$, 
  Huimeng Wang$^1$, 
  Shuhai Peng$^3$, \\
  {\bf Shiyin Kang}$^{2}$\thanks{Corresponding author} \bf ,
  {\bf Xixin Wu}$^1$\bf ,
  {\bf Zhiyong Wu}$^3$ \\
  $^1$The Chinese University of Hong Kong,
  $^2$SenseTime Research, 
  $^3$Tsinghua University \\
  \texttt{\{luhui, xychen, huimengwang, wuxx, zywu\}@se.cuhk.edu.hk} \\
  \texttt{psh24@mails.tsinghua.edu.cn, kangshiyin@sensetime.com}
}
\begin{document}

\maketitle

\begin{abstract}
Full-duplex spoken dialogue requires a model to keep listening while generating its own spoken response. This is challenging for large language models (LLMs), which are designed to extend a single coherent sequence and do not naturally support user input arriving during generation. We argue that how the user stream is routed into the LLM is therefore a key architectural question for full-duplex modeling. To study this question, we extend a text-only LLM into a unified full-duplex spoken dialogue system and compare two routing strategies under a shared training pipeline: (i) channel fusion, which injects the user stream directly into the LLM input, and (ii) cross-attention routing, which keeps the user stream as external memory accessed through cross-attention adapters.
Experiments on spoken question answering and full-duplex interaction benchmarks reveal a clear tradeoff. Channel fusion yields stronger semantic grounding and consistently better question-answering performance. However, under semantically overlapping conditions such as user interruptions, it is more vulnerable to context corruption: if the model fails to stop in time, the overlapping user stream can interfere with ongoing generation and lead to semantically incoherent continuations. Cross-attention routing underperforms on question answering, but better preserves the LLM generation context and is more robust to this failure mode. These results establish user-stream routing as a central design axis in full-duplex spoken dialogue and offer practical guidance on the tradeoff between semantic integration and context robustness. We provide a demo page\footnote{https://light1726.github.io/duplex-demo/} for qualitative inspection.
\end{abstract}

\section{Introduction}
\label{intro}
Speech is one of the most natural modalities for human communication, making spoken dialogue a key interface for human--machine interaction. Recent advances in large language models (LLMs) \citep{DBLP:conf/nips/VaswaniSPUJGKP17,NEURIPS2020_1457c0d6} have made them an increasingly common foundation for spoken dialogue systems.
Most existing spoken dialogue models adopt a strict turn-taking assumption. Under this formulation, the user and the system speak in alternation, without overlapping speech, so the interaction can be formulated into a single interleaved sequence of utterances. This corresponds to a half-duplex communication pattern, in which only one party is effectively transmitting information at a time. Many recent LLM-based spoken dialogue systems follow this setup: the model consumes the user’s complete speech utterance as input and then autoregressively generates a spoken response \citep{DBLP:journals/corr/abs-2407-10759,DBLP:journals/corr/abs-2412-02612,DBLP:conf/iclr/FangGZMZ025,fang-etal-2025-llama,DBLP:journals/corr/abs-2502-17239,DBLP:journals/corr/abs-2504-18425,DBLP:journals/corr/abs-2507-16632}.

However, real-world spoken dialogue is often not strictly turn-based \citep{360380fc-f5d2-336c-aecb-167b93802825}. Speakers may interrupt one another or produce short backchannels while the other party is still speaking. In such cases, the two audio streams overlap in time. This is often referred to as the full-duplex spoken dialogue, where both parties can listen and speak simultaneously. Modeling full-duplex dialogue is important for more natural and fluid voice interaction, as highlighted by GPT-4o \citep{DBLP:journals/corr/abs-2410-21276} and a growing body of recent work on full-duplex spoken dialogue systems \citep{DBLP:journals/tacl/NguyenKCAHETASM23,DBLP:journals/corr/abs-2410-00037,zhang-etal-2024-beyond,veluri-etal-2024-beyond,zhang-etal-2025-omniflatten,DBLP:conf/interspeech/HuHCCGZCLBG25,DBLP:journals/corr/abs-2505-17060}.
% \begin{figure}[t]
% \centering

% \begin{subfigure}{0.49\columnwidth}
% \centering
% \includegraphics[width=\linewidth]{images/strict_turn_taking.pdf}
% \caption{Strict turn-taking spoken dialogue}
% \label{fig:strict_turn_taking}
% \end{subfigure}
% \hfill
% \begin{subfigure}{0.49\columnwidth}
% \centering
% \includegraphics[width=\linewidth]{images/full_duplex.pdf}
% \caption{Full-duplex spoken dialogue}
% \label{fig:full_duplex}
% \end{subfigure}

% \caption{Illustration of strict turn-taking and full-duplex spoken dialogue}
% \label{fig:self_attention_flow}
% \end{figure}

Nevertheless, the dual-stream nature of full-duplex spoken dialogue is not naturally aligned with how LLMs are pre-trained. Standard LLMs are optimized to extend a single-stream token sequence through next-token prediction. In full-duplex dialogue, however, the model must continue generating its own response while also processing an external user speech stream to decide whether to keep generating, or stop speaking. This raises a central architectural question that prior work has largely addressed only implicitly: \emph{how should the user stream be routed into the LLM during generation?} The answer is important because it determines how incoming user speech interacts with the model’s ongoing generation state, and therefore how robustly the model can behave under overlapping speech.

Existing full-duplex systems mainly address this question in two ways. The first is to segment both audio streams into fixed-size chunks and interleave them into a single-stream sequence \citep{veluri-etal-2024-beyond,DBLP:journals/corr/abs-2501-06282,zhang-etal-2025-omniflatten,DBLP:journals/corr/abs-2505-17060}. The LLM then models full-duplex dialogue as autoregressive prediction over this interleaved sequence. While conceptually simple, interleaving effectively doubles the sequence length and often inserts many silent user chunks, leading to additional KV-cache and computation overhead.

The second strategy is channel fusion, which combines the two streams at each time step before feeding them into the LLM \citep{DBLP:conf/interspeech/HuHCCGZCLBG25,DBLP:journals/corr/abs-2509-02521,DBLP:journals/corr/abs-2512-20156}. Channel fusion is compact, since the sequence length remains the same as that of a single audio stream, and it gives the LLM direct, time-aligned access to the user stream at every step. However, once the two streams are fused, the model no longer has an explicit architectural mechanism to separate user input from its own generation context. This is particularly problematic when the two streams overlap in semantically meaningful ways, such as during user interruptions. If the model fails to stop promptly, the overlapping user speech may interfere with the context driving continued generation, leading to corrupted or semantically incoherent continuations. To our knowledge, this failure mode has not been systematically characterized in prior work.

Motivated by this limitation, we revisit a structurally separated alternative: routing the user stream through cross-attention adapters \citep{DBLP:conf/nips/AlayracDLMBHLMM22}. Under this design, the user stream is represented as an external memory of keys and values, while the LLM’s own generation remains in its native autoregressive context. This provides an explicit and gateable mechanism for attending to user speech without forcing it into the same context used for token generation. Although cross-attention adapters have been used for visual conditioning \citep{DBLP:conf/nips/AlayracDLMBHLMM22} and audio understanding \citep{DBLP:conf/icml/KongGBPVC24}, their role in full-duplex spoken dialogue, and their tradeoff relative to direct channel fusion, has not been systematically studied.

In this paper, we develop a unified framework for extending a text-only LLM into a full-duplex spoken dialogue model through task formulation, tailored data construction, and a staged training curriculum, including practical designs for user interruption handling. This framework provides a controlled testbed for studying user-stream routing as a central architectural design axis in full-duplex spoken dialogue. Within this shared setup, we compare two model variants under matched model and training conditions: \textbf{CF-Duplex}, which uses channel fusion, and \textbf{XA-Duplex}, which uses cross-attention routing.
Our results reveal a clear tradeoff. \textbf{CF-Duplex} yields stronger spoken question answering performance, suggesting better semantic grounding from speech input. However, during user interruptions, the same direct conditioning can become a liability: when the model fails to stop in time, \textbf{CF-Duplex} is more prone to producing semantically incoherent continued generation, whereas \textbf{XA-Duplex} largely avoids this failure mode. We further examine this behavior through qualitative analysis of failed interruption cases, helping clarify when each routing strategy is preferable.

Our contributions are threefold: (i) we develop a unified framework for extending a text-only LLM into a full-duplex spoken dialogue model through task formulation, data construction, and a staged training curriculum, and introduce practical designs for user interruption handling; (ii) we identify user-stream routing as a key architectural design question in full-duplex spoken dialogue; and (iii) within a controlled experimental setting based on a shared system design and training curriculum, we compare channel fusion and cross-attention routing across automatic speech recognition (ASR), text-to-speech synthesis (TTS), turn-based spoken dialogue, and full-duplex spoken dialogue, revealing a tradeoff between stronger semantic integration and greater context robustness.

\section{Related works}
\subsection{Speech-based LLMs and half-duplex spoken dialogue models}
Recent work has increasingly adopted LLMs as a unified backbone for speech-language processing, including ASR, TTS, audio understanding, and spoken dialogue \citep{zhang-etal-2023-speechgpt,DBLP:journals/corr/abs-2407-10759,DBLP:journals/corr/abs-2412-02612,DBLP:journals/corr/abs-2503-20215,DBLP:journals/corr/abs-2502-17239,DBLP:journals/corr/abs-2504-18425,DBLP:conf/icml/KongGBPVC24,ghosh2025audio,ghosh2026audio}. A common paradigm is to map speech into the LLM embedding space and represent speech and text as a single sequence that can be processed autoregressively. Recent systems further enable end-to-end spoken dialogue by generating both text and speech outputs within a unified framework \citep{DBLP:journals/corr/abs-2412-02612,DBLP:conf/iclr/FangGZMZ025,fang-etal-2025-llama,tan2026drvoice}.
Despite their strong performance, these models typically assume strict turn taking: the system first consumes the user’s complete utterance and only then generates its response. As a result, they do not naturally support full-duplex behaviors such as interruption and backchannels handling. Our work builds on this line of speech-based LLMs, but extends it to full-duplex spoken dialogue through streaming speech processing adaptation, data construction, task design, and training.
\subsection{Full-duplex spoken dialogue models}
Full-duplex spoken dialogue can be supported either by coordinating an LLM with external modules such as voice activity or turn detection \citep{DBLP:conf/emnlp/ZhangLMKPTTY23,DBLP:journals/corr/abs-2509-23938,DBLP:conf/icml/WangLFZS000M25}, or by adapting the LLM itself to model overlapping user and system speech end to end. While practical, external modules add latency and are not jointly optimized with the LLM.
Recent end-to-end approaches mainly differ in how they represent and route the user and system streams during generation. One line of work interleaves the two streams into a single-stream sequence, as in SyncLLM \citep{veluri-etal-2024-beyond}, OmniFlatten \citep{zhang-etal-2025-omniflatten}, NTPP \citep{DBLP:conf/icml/WangMCZWWKCZ25}, and SALMONN-omni \citep{DBLP:journals/corr/abs-2505-17060}. This yields a simple autoregressive formulation, but increases sequence length and often introduces many silent user chunks into the context. Another line adopts channel fusion, combining the user and system streams at each time step before entering the LLM, as in Moshi \citep{DBLP:journals/corr/abs-2410-00037}, LSLM \citep{10.1609/aaai.v39i23.34665}, SLAM-duplex \citep{DBLP:conf/interspeech/HuHCCGZCLBG25}, FLM-Audio \citep{DBLP:journals/corr/abs-2509-02521}, and Fun-Audio-Chat \citep{DBLP:journals/corr/abs-2512-20156}.
Other architectures have also been explored, including dual decoders with cross-stream attention \citep{DBLP:journals/tacl/NguyenKCAHETASM23}, dual-LLM designs \citep{DBLP:journals/corr/abs-2408-05211}, and encoder--decoder models that jointly encode both streams before decoding \citep{DBLP:journals/corr/abs-2501-04877}. In contrast, our cross-attention variant keeps the user stream as a separate memory accessed during generation. To our knowledge, this form of separated user-stream routing has not been systematically compared with channel fusion under a shared modeling and training setup for full-duplex spoken dialogue.

\subsection{LLMs with cross-attention conditioning}
Cross-attention has been widely adopted in multimodal LLMs as a mechanism for conditioning a pretrained text-only LLM on external inputs. Flamingo \citep{DBLP:conf/nips/AlayracDLMBHLMM22} introduces cross-attention adapters for visual conditioning, and AudioFlamingo \citep{DBLP:conf/icml/KongGBPVC24} extends this design to audio understanding. Related ideas have also been explored in streaming video understanding and reasoning, where cross-attention is used to route visual input into the LLM while preserving a separate textual generation context \citep{DBLP:journals/corr/abs-2412-08646}.
These works motivate cross-attention as a plausible mechanism for conditioning LLMs on external streaming input. However, its role in full-duplex spoken dialogue remains underexplored, particularly as a user-stream routing strategy to be compared directly with channel fusion.
\begin{figure}
    \centering
    \includegraphics[width=0.98\linewidth]{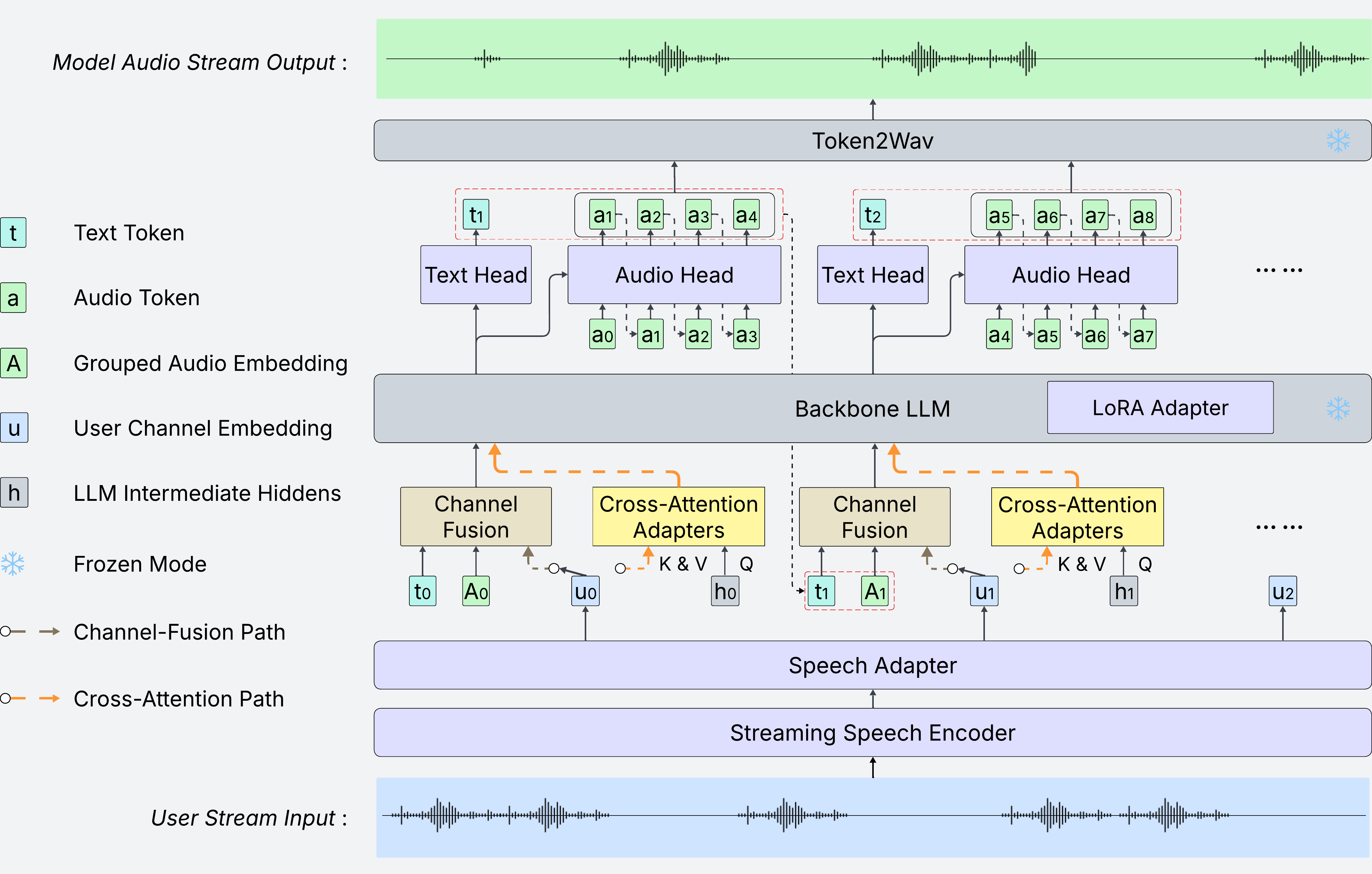}
    \caption{Model architecture}
    \label{fig:architecture}
\end{figure}

\section{Architecture}
\label{method}
We extend a pretrained text-only LLM into a full-duplex spoken dialogue system with streaming speech input and output. The system comprises a streaming speech encoder and adapter for user audio, a text-only LLM backbone, an audio decoder for spoken response generation, and a user-stream routing module. We study two modeling variants, \textbf{CF-Duplex} and \textbf{XA-Duplex}, based on channel fusion and cross-attention routing strategies, respectively. Figure~\ref{fig:architecture} illustrates the overall architecture.
We describe the shared modules and the two routing variants in the following subsections.

\subsection{Speech encoder and adapter}

A full-duplex spoken dialogue system must process user audio incrementally so that it can decide, in real time, whether to continue listening, begin responding, or stop speaking. To this end, we initialize the speech encoder from pretrained Whisper \citep{DBLP:conf/icml/RadfordKXBMS23} and adapt it for streaming speech processing. We segment the input waveform into fixed-size chunks, extract mel-spectrogram features for each chunk independently, and feed the resulting spectrogram chunks to the encoder incrementally. During inference, the encoder maintains continuity across chunks using KV caching.

The original Whisper encoder consists of convolutional layers followed by bidirectional self-attention blocks. To preserve causality in streaming operation, we left-pad the convolutional input and apply causal masks to all self-attention layers. In addition, the original Whisper encoder uses sinusoidal positional embeddings, which are less suitable for long-form incremental processing. We therefore replace them with RoPE \citep{DBLP:journals/ijon/SuALPBL24} and finetune them jointly with the adapted encoder.

We then apply a speech adapter to map the encoder output to the representation space expected by the backbone LLM. Specifically, the adapter consists of three linear layers, with the middle layer operating on pairs of consecutive frames concatenated along the feature dimension. This reduces the time resolution of the user stream by a factor of 2.

\subsection{Speech tokenization and audio head}
\label{subsec:speech_tok_audio_head}

We use the supervised semantic speech tokenizer from CosyVoice~2 \citep{DBLP:journals/corr/abs-2412-10117} to convert speech waveforms into discrete tokens at 25\,Hz, and use the pretrained \texttt{token2wav} model from CosyVoice~2 for waveform reconstruction.
On the output side, system speech is represented as discrete audio tokens and generated by an autoregressive audio decoder, which we refer to as the audio head. The audio head is implemented as a lightweight decoder LLM conditioned on last-layer hidden states from the backbone LLM. We choose this design instead of a shallow projection layer because mapping backbone representations to audio-token sequences is substantially more complex than standard text-token prediction.
Because the audio-token sequence is much longer than the corresponding text-token sequence, we follow recent work \citep{tan2026drvoice,DBLP:journals/corr/abs-2512-20156} and generate groups of \(\mathcal{G}\) consecutive audio tokens conditioned on the same backbone hidden state. This reduces the rate mismatch between text and audio generation. During decoding, each generated audio-token group is embedded and fed back into the backbone as the model-audio stream.
To preserve a roughly causal relationship between text and speech generation, we introduce a delay factor \(\mathcal{D}\), so that audio decoding starts only after \(\mathcal{D}\) text tokens have been generated. During training, we pad the shorter sequence at the end so that the text-token sequence and the audio-token-group sequence have the same length.

\subsection{User-stream routing}
We compare two strategies for routing the user stream into the backbone LLM: channel fusion and cross-attention routing, yielding two model variants, \textbf{CF-Duplex} and \textbf{XA-Duplex}, respectively.
The user stream, model text stream, and model audio stream are defined on a shared duplex timeline, so aligned positions correspond to the same underlying interaction time step.
For CF-Duplex, we directly fuse the user-stream embeddings with the model text and audio streams at each aligned time step, and the backbone LLM operates on the resulting sequence as a single autoregressive stream. Let the user stream, model text stream, and model audio stream be denoted by \(\mathbf{u}\), \(\mathbf{m}_{\mathrm{text}}\), and \(\mathbf{m}_{\mathrm{audio}}\), respectively. Let \(\mathbf{c}=[\mathbf{u};\mathbf{m}_{\mathrm{text}};\mathbf{m}_{\mathrm{audio}}]\) denote concatenation along the feature dimension. The fusion process is defined as
\begin{equation}
    \label{eqn:channel_fusion}
    \mathbf{y}
    = \mathbf{u}+\mathbf{m}_{\mathrm{text}}+\mathbf{m}_{\mathrm{audio}}
    + \boldsymbol{\sigma}\!\big(W_g\mathbf{c}+\mathbf{b}_g\big)
    \odot \mathrm{MLP}(\mathbf{c}),
\end{equation}
where \(\boldsymbol{\sigma}\) is the element-wise sigmoid gate and \(\mathrm{MLP}\) is a two-layer multilayer perceptron. This design gives the backbone LLM direct access to time-aligned user stream at every step, while also merging the user stream into the same context used for the system's own generation.

For XA-Duplex, we keep the user stream as a separate memory and let the LLM access it through a set of cross-attention adapters. The user-stream embeddings serve as keys and values, while the intermediate hidden states of the backbone LLM serve as queries. We adopt the XA-Dense variant used in Flamingo \citep{DBLP:conf/nips/AlayracDLMBHLMM22} and AudioFlamingo \citep{DBLP:conf/icml/KongGBPVC24}. To preserve temporal correspondence across streams, we assign the user stream the same timeline indices as the model streams and apply RoPE accordingly before feeding it into the cross-attention adapters. This allows intermediate LLM layers to attend to temporally aligned user evidence while keeping the user stream separate from the backbone LLM's native autoregressive generation context.
\section{Training strategy}
\label{sec:training_strategy}

\begin{figure}
    \centering
    \includegraphics[width=0.98\linewidth]{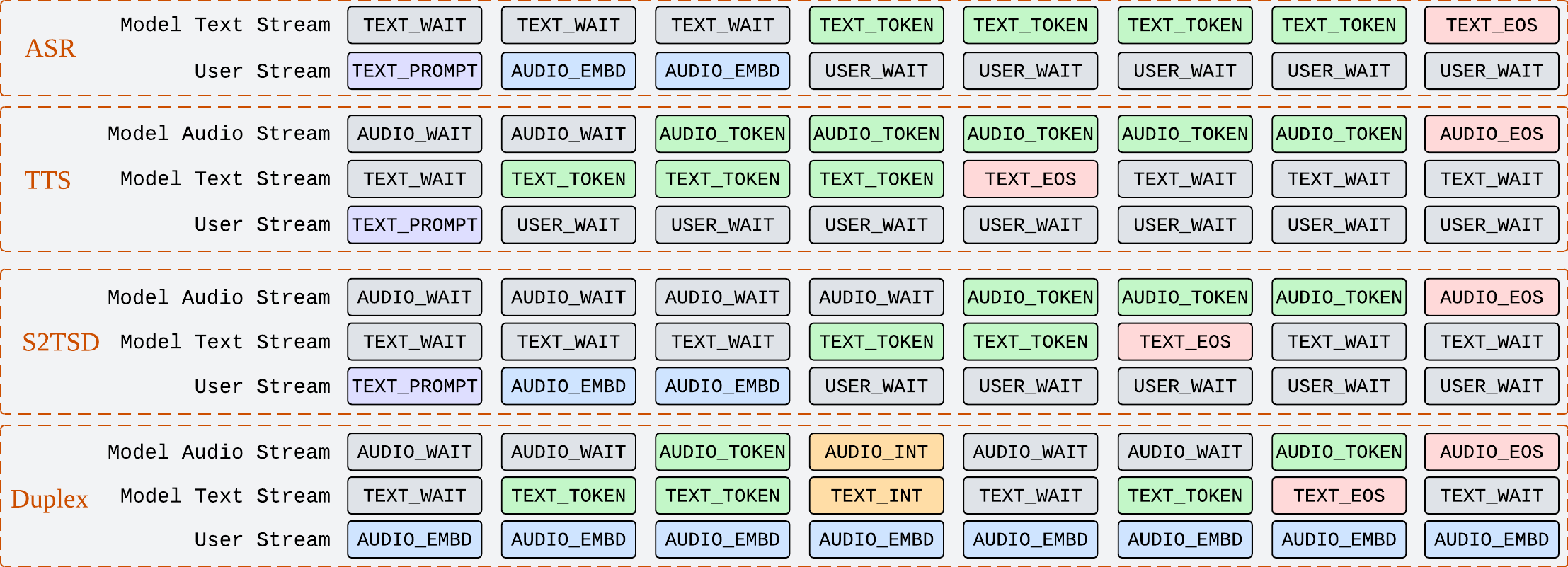}
    \caption{Input and output data formats for different tasks}
    \label{fig:data_format}
\end{figure}

To train the full-duplex spoken dialogue system, we use a staged curriculum spanning ASR, streaming TTS, speech-to-text dialogue (S2TD), speech-to-text-and-speech dialogue (S2TSD), and full-duplex spoken dialogue. We formulate all tasks using a unified stream-based format compatible with the target full-duplex setting, as illustrated in Figure~\ref{fig:data_format}. Each example is represented using a user stream, a model text stream, and, when applicable, a model audio stream.

\subsection{Task formulation and sequence construction}

We use special tokens to explicitly represent idle and interruption behavior along the duplex timeline. On the user side, \texttt{<USER\_WAIT>} denotes intervals in which the user is silent while waiting for the system response. On the model side, \texttt{<TEXT\_WAIT>} and \texttt{<AUDIO\_WAIT>} denote intervals in which the model has not yet started responding or has finished responding and is waiting for further user input. For duplex dialogue, we additionally use \texttt{<TEXT\_INT>} and \texttt{<AUDIO\_INT>} to provide explicit supervision for interruption handling. We treat these special tokens as ordinary prediction targets in the corresponding streams, so that waiting and interruption behavior are learned through explicit token-level supervision. The details regarding how input and output sequences in different tasks are composed are described in Appendix \ref{appendix:task_formulation}.

\subsection{Training curriculum}

We train both CF-Duplex and XA-Duplex using the same three-stage curriculum, progressing from basic speech perception and generation to turn-based dialogue and finally to full-duplex interaction.
In Stage~1, we train the speech encoder, speech adapter, routing-specific modules, audio head, and LoRA adapters \citep{DBLP:conf/iclr/HuSWALWWC22} of the backbone LLM on ASR and streaming TTS. In Stage~2, we freeze the speech encoder and backbone LLM and train the remaining modules on ASR, streaming TTS, S2TD, and S2TSD. In Stage~3, we train the same set of modules as in stage~2 on ASR, streaming TTS and full-duplex spoken dialogue, where the user input is a continuous audio stream containing silence, queries, interruptions, and backchanneling. Using the same curriculum for both variants helps isolate the effect of the routing design.

\section{Data preparation and construction}
\label{sec:data_construction}

We prepare training data for ASR, TTS, turn-based spoken dialogue, and full-duplex spoken dialogue. Detailed dataset statistics, prompts, and construction examples are provided in the appendix~\ref{appendix:data_construction_details}.

\noindent\textbf{ASR and TTS data.}
For ASR, we combine several openly available English speech corpora, including LibriSpeech \citep{DBLP:conf/icassp/PanayotovCPK15}, GigaSpeech \citep{DBLP:conf/interspeech/ChenCWDZWSPTZJK21}, PeopleSpeech \citep{NEURIPS_DATASETS_AND_BENCHMARKS2021_202cb962}, MLS \citep{DBLP:conf/interspeech/PratapXSSC20}, CommonVoice \citep{ardila-etal-2020-common}, VoxPopuli \citep{wang-etal-2021-voxpopuli}, and Emilia-Large \citep{DBLP:journals/corr/abs-2501-15907}, yielding 217k hours of English ASR data in total. For TTS, we use VoxBox \citep{DBLP:journals/corr/abs-2503-01710}, which provides 104k hours of English and Chinese speech synthesis data.

\noindent\textbf{Turn-based spoken dialogue.}
To construct turn-based spoken dialogue data for S2TD and S2TSD training, we start from publicly available textual QA and dialogue datasets, including SQuAD \citep{rajpurkar-etal-2016-squad}, MS-MARCO \citep{DBLP:conf/nips/NguyenRSGTMD16}, HotpotQA \citep{yang-etal-2018-hotpotqa}, Natural Questions \citep{kwiatkowski-etal-2019-natural}, UltraChat \citep{ding-etal-2023-enhancing}, and \texttt{I\_Wonder\_Why-Chinese} \footnote{https://huggingface.co/datasets/Mxode/I\_Wonder\_Why-Chinese}. We use \texttt{Qwen3-30B-A3B-Instruct-2507} \footnote{https://huggingface.co/Qwen/Qwen3-30B-A3B-Instruct-2507} to rewrite these samples into English spoken-style question--answer pairs and then synthesize them into speech using IndexTTS-2 \citep{DBLP:conf/aaai/ZhouZHZWDS26}. We further filter out samples with overly long audio or poor text--audio consistency. After filtering, we obtain 1.9M turn-based spoken dialogue samples.

\noindent\textbf{Full-duplex spoken dialogue.}
We construct full-duplex spoken dialogue by composing turn-based spoken dialogues into two-channel audio interactions. Based on these base conversations, we simulate two common full-duplex behaviors: user interruptions and backchanneling. For interruptions, we construct both context-dependent and context-independent cases. In the context-dependent case, we use \texttt{Qwen3-30B-A3B-Instruct-2507} to generate a follow-up question triggered by a phrase in the initial response, and only allow the interruption to occur after that trigger phrase during the spoken dialogue construction. In the context-independent case, we insert a question--answer pair sampled from another conversation. For backchannels, we collect a list of user backchanneling words and append one into a base conversation.
We construct full-duplex examples on the fly during training rather than pre-composing fixed two-channel conversations. This allows us to randomize insertion timing for interruptions and backchannels while enforcing semantic constraints for context-dependent interruptions. We also vary the model's reaction delay to interruptions during training, which improves interruption handling in our ablations (see Section \ref{sec:exp:ablation}).

\section{Experiments}
\label{sec:exp}
\subsection{Modeling and training configuration}
We use \texttt{Qwen3-1.7B} \footnote{https://huggingface.co/Qwen/Qwen3-1.7B} as the backbone LLM and keep it frozen throughout training. We train LoRA adapters \citep{DBLP:conf/iclr/HuSWALWWC22} on top of the backbone, with rank 16 and scaling factor \(\alpha=32\). The audio head is initialized from \texttt{Qwen3-0.6B} \footnote{https://huggingface.co/Qwen/Qwen3-0.6B}.
For speech generation, we set the audio grouping size to \(\mathcal{G}=4\) and the delay factor to \(\mathcal{D}=2\). We find \(\mathcal{G}=4\) to work better than \(\mathcal{G}=5\) (which is adopted in \citep{DBLP:journals/corr/abs-2512-20156}) in stage-1 training; the comparison is provided in Appendix~\ref{appendix:grouping_size}.
For the XA-Duplex, we insert cross-attention layers into the even-numbered layers of the backbone LLM. We select this placement based on stage-1 ASR and TTS performance; details are provided in Appendix~\ref{appendix:xa_positions}. Other modeling and training specifications of our models can be found in Appendix~\ref{appendix:training_specs}.

% We train all stages on 16 NVIDIA H200 GPUs using AdamW \citep{DBLP:conf/iclr/LoshchilovH19} with \(\beta_1=0.9\), \(\beta_2=0.999\), and weight decay 0.01. We use a learning rate of \(1\times10^{-4}\) for Stage~1, \(5\times10^{-5}\) for Stage~2, and \(3\times10^{-5}\) for Stage~3, with linear warm-up for 8k, 4k, and 4k steps, respectively, followed by cosine decay. We use dynamic batching with an upper limit on the total number of text tokens or audio-token groups per batch: 4,800 for Stage~1, 3,600 for Stage~2, and 2,400 for Stage~3.

\subsection{Metrics}
\label{sec:metrics}
We evaluate the proposed models on ASR, TTS, S2TD, S2TSD, and full-duplex spoken dialogue. For ASR, we report word error rate (WER) on LibriSpeech \citep{DBLP:conf/icassp/PanayotovCPK15} \texttt{test-clean} and \texttt{test-other}. For TTS, we evaluate on \texttt{seed-tts-eval}\footnote{https://github.com/BytedanceSpeech/seed-tts-eval} and report WER on its English and Chinese subsets.
For spoken question answering, we use LLaMA Questions (LLaMAQ)\footnote{https://github.com/google-research-datasets/LLAMA1-Test-Set}, TriviaQA (TriviaQ) \citep{DBLP:conf/acl/JoshiCWZ17}, WebQuestions (WebQ) \footnote{https://huggingface.co/datasets/stanfordnlp/web\_questions} and AlpacaEval \citep{alpaca_eval}, with audio questions from OpenAudioBench\footnote{https://huggingface.co/datasets/baichuan-inc/OpenAudioBench}. Following the OpenAudioBench protocol, we use its evaluation prompts with \texttt{GPT-5.4-mini}\footnote{https://openai.com/index/introducing-gpt-5-4-mini-and-nano/} to judge answer correctness and quality for all four datasets. For S2TSD and full-duplex spoken dialogue, we evaluate both textual and spoken responses.
For full-duplex interaction behavior, we report results on \texttt{Full-Duplex-Bench v1.0 \& v1.5} \citep{DBLP:conf/asru/LinLLWALL25,lin2025fdb_v15}. On \texttt{v1.0}, we evaluate User Interruption and Smooth Turn Taking using takeover rate (TOR), response quality in GPT-4o score, and response latency. On \texttt{v1.5}, we evaluate User Interruption and User Backchannel using the benchmark-defined behavior scores, stop latency, and response latency. Detailed definitions are provided in Appendix~\ref{appendix:full_duplex_metrics} and can be referred to from the original papers.

\subsection{Results}
\subsubsection{Question answering}
We compare both CF-Duplex and XA-Duplex with representative half-duplex and full-duplex spoken dialogue models. Table \ref{tab:qa} demonstrates that CF-Duplex achieves competitive question answering performance despite using a much smaller backbone LLM (\texttt{Qwen3-1.7B}) than most baselines. In particular, it remains comparable to prior full-duplex systems on LLaMAQ and TirviaQ, achieves the best spoken response quality on WebQ, and attains the best performance on AlpacaEval among all compared models. These results suggest that the proposed CF-Duplex can effectively support spoken dialogue understanding even under a compact model scale and training budget.

At the same time, Table \ref{tab:qa} also reveals a clear gap between our two model variants: XA-Duplex consistently underperforms CF-Duplex across all four datasets in both speech and text responses. This indicates that, under the same backbone LLM, the CF-Duplex architecture provides substantially stronger QA capability than the XA-Duplex.
\begin{table}[ht]
  \caption{Question answering performance (speech | text score)}
  \label{tab:qa}
  \centering
  \resizebox{\columnwidth}{!}{
  \begin{tabular}{lcccccc}
    \toprule
    Method        & Backbone LLM & Full-Duplex& LLaMAQ  & TriviaQ  & WebQ  & AlpacaEval \\
    \midrule
    Speech-GPT \citep{zhang-etal-2023-speechgpt}    & \texttt{LLaMA-13B}      & \xmark & - | 21.0    & - | 14.8    & - | 6.5 & -\\ 
    GLM-4-Voice \citep{DBLP:journals/corr/abs-2412-02612}   & \texttt{GLM-4-9B-Base} & \xmark & 50.7 | 64.7 & 26.5 | 39.1 & 15.9 | 32.2 & 3.58 | 3.82 \\
    \midrule
    Freeze-Omni \citep{DBLP:conf/icml/WangLFZS000M25}  & \texttt{Qwen2-7B-Instruct}  &\cmark  & \textbf{56.2} | \textbf{74.2} & \textbf{28.5} | \textbf{45.1} & 27.9 | \textbf{40.8} & 2.46 | 3.90 \\
    Moshi \citep{DBLP:journals/corr/abs-2410-00037}        & \texttt{Helium (7B)} &\cmark  & 54.5 | 60.8 & 16.7 | 25.6 & 22.1 | 23.4 & 1.76 | 1.84 \\
    SALM-Duplex \citep{DBLP:conf/interspeech/HuHCCGZCLBG25}  & \texttt{TinyLlama-1.1B-chat} &\cmark & 51.3 | - & 16.9 | - & 25.0 | - & 2.99 | - \\
    \midrule
    CF-Duplex & \texttt{Qwen3-1.7B}    & \cmark & 50.7 | 57.3 & 18.1 | 19.6 & \textbf{28.0} | 30.3 & \textbf{3.94} | \textbf{4.16} \\
    XA-Duplex & \texttt{Qwen3-1.7B}   & \cmark & 38.3 | 40.3 & 8.0 | 8.2   & 18.4 | 18.7 & 3.87 | 4.04 \\
    \bottomrule
  \end{tabular}}
\end{table}
% \begin{table}[ht]
%   \caption{Question answering performance (Speech | Text accuracy).}
%   \label{tab:qa}
%   \centering
%   \resizebox{\columnwidth}{!}{
%   \begin{tabular}{lccccc}
%     \toprule
%     Method        & Backbone LLM & Full-Duplex& LLaMAQ (S | T) & TriviaQ (S | T) & WebQ (S | T) \\
%     \midrule
%     Speech-GPT \citep{zhang-etal-2023-speechgpt}    & \texttt{LLaMA-13B}      & \xmark & - | 21.0    & - | 14.8    & - | 6.5 \\ 
%     GLM-4-Voice \citep{DBLP:journals/corr/abs-2412-02612}   & \texttt{GLM-4-9B-Base} & \xmark & 50.7 | 64.7 & 26.5 | 39.1 & 15.9 | 32.2\\
%     \midrule
%     Freeze-Omni \citep{DBLP:conf/icml/WangLFZS000M25}  & \texttt{Qwen2-7B-Instruct}  &\cmark  & \textbf{56.2} | \textbf{74.2} & \textbf{28.5} | \textbf{45.1} & 27.9 | \textbf{40.8} \\
%     Moshi \citep{DBLP:journals/corr/abs-2410-00037}        & \texttt{Helium (7B)} &\cmark  & 54.5 | 60.8 & 16.7 | 25.6 & 22.1 | 23.4 \\
%     SALM-Duplex \citep{DBLP:conf/interspeech/HuHCCGZCLBG25}  & \texttt{TinyLlama-1.1B-chat} &\cmark & 51.3 | - & 16.9 | - & 25.0 | - \\
%     \midrule
%     CF-Duplex & \texttt{Qwen3-1.7B}    & \cmark & 50.7 | 57.3 & 18.1 | 19.6 & \textbf{28.0} | 30.3  \\
%     XA-Duplex & \texttt{Qwen3-1.7B}   & \cmark & 38.3 | 40.3 & 8.0 | 8.2   & 18.4 | 18.7  \\
%     \bottomrule
%   \end{tabular}}
% \end{table}
\subsubsection{Full-duplex behavior}
We further evaluate interaction behavior on \texttt{Full-Duplex Bench v1.0} and \texttt{v1.5}. As shown in Tables \ref{tab:fdbench-v1} and \ref{tab:fdbench-v1_5}, CF-Duplex delivers the strongest overall performance among our model variants and remains highly competitive with prior full-duplex systems. On Full-Duplex Bench v1.0, CF-Duplex achieves a perfect takeover rate in the User Interruption scenario (TOR = 1.000), matching the best reported result, while obtaining the highest response quality score (GPT-4o = 3.96). Although its interruption latency is slightly higher than that of the fastest systems, it remains low overall. In Smooth Turn Taking, CF-Duplex also attains strong performance, with high TOR and low latency, indicating reliable turn-end detection and prompt response generation.
\begin{table}[htbp]
\centering
\caption{Results on \texttt{Full-Duplex Bench v1.0}}
\label{tab:fdbench-v1}
\resizebox{0.7\columnwidth}{!}{
\begin{tabular}{l c c c c c}
\toprule
Models & \multicolumn{3}{c}{User Interruption} & \multicolumn{2}{c}{Smooth Turn Taking} \\
\cmidrule(lr){2-4} \cmidrule(lr){5-6}
& TOR (\(\uparrow\)) & GPT-4o (\(\uparrow\)) & Latency (\(\downarrow\)) & TOR (\(\uparrow\)) & Latency (\(\downarrow\)) \\
\midrule
dGSLM \citep{DBLP:journals/tacl/NguyenKCAHETASM23}   & 0.917 & 0.201 & 2.531 & 0.975 & 0.352 \\
Freeze-Omni \citep{DBLP:conf/icml/WangLFZS000M25}    & 0.867 & 3.615 & 1.409 & 0.336 & 0.953 \\
Moshi \citep{DBLP:journals/corr/abs-2410-00037}      & \textbf{1.000} & 0.765 & \textbf{0.257} & 0.941 & 0.265 \\
Gemini Live \citep{DBLP:conf/asru/LinLLWALL25} & 0.891 & 3.376 & 1.183 & 0.655 & 1.301 \\
\midrule
CF-Duplex  & \textbf{1.000} & \textbf{3.96} & 0.374 & 0.924 & 0.336 \\
XA-Duplex & 0.971 & 2.23 & 0.325 & \textbf{0.983} & \textbf{0.161} \\
\bottomrule
\end{tabular}}
\end{table}

On \texttt{Full-Duplex Bench v1.5}, CF-Duplex shows particularly strong control over user interruptions. In the Interruption scenario, it achieves the highest or tied-highest \texttt{Respond} score (0.72), while also obtaining the lowest stop latency (0.736 s) and the lowest respond latency (0.724 s) among all compared models. In the Backchannel scenario, it achieves the highest \texttt{Resume} rate (0.96). These results indicate that CF-Duplex can effectively tackle with user interruptions and backchannels.

In contrast, while XA-Duplex performs competitively on some timing-related metrics, such as backchannel response latency, and achieves the best result on Full-duplex \texttt{v1.0} Smooth Turn-Taking, its overall behavior is less robust. We also observe that it sometimes responds too aggressively by talking over the user. In particular, it underperforms CF-Duplex on interruption handling, with substantially lower response quality on \texttt{v1.0} and a much lower \texttt{Respond} score on \texttt{v1.5}. Overall, these results suggest that CF-Duplex provides a better balance between responsiveness, behavioral accuracy, and conversational quality.
% In contrast, while XA-Duplex performs competitively on some timing-related metrics, such as smooth turn-taking latency and backchannel response latency, its overall behavior is less robust. In particular, it underperforms CF-Duplex on interruption handling, with substantially lower response quality on \texttt{v1.0} and a much lower \texttt{Respond} score on \texttt{v1.5}. Overall, these results suggest that CF-Duplex provides a better balance between responsiveness, behavioral accuracy, and conversational quality.

\begin{table}[ht]
  \caption{Results on \texttt{Full-Duplex Bench v1.5}}
  \label{tab:fdbench-v1_5}
  \centering
  \resizebox{0.9\columnwidth}{!}{
  \begin{tabular}{llcccc}
    \toprule
    Task & Metric & Freeze-Omni \citep{DBLP:conf/icml/WangLFZS000M25} & Moshi \citep{DBLP:journals/corr/abs-2410-00037} & CF-Duplex & XA-Duplex \\
    \midrule
    \multirow{6}{*}{Interruption}
      & Respond ($\uparrow$)           & \textbf{0.72} & 0.50 & \textbf{0.72} & 0.32 \\
      & Resume ($\downarrow$)          & \textbf{0.12} & 0.26 & 0.17 & 0.41 \\
      & Uncertain ($\downarrow$)       & 0.03 & \textbf{0.00} & 0.04 & 0.05 \\
      & Unknown ($\downarrow$)         & 0.13 & 0.25 & \textbf{0.07} & 0.23 \\
      \cmidrule(r){2-6}
      & Stop Latency ($\downarrow$)    & 1.42 & 1.16 & \textbf{0.74} & 1.18 \\
      & Respond Latency ($\downarrow$) & 1.35 & 1.47 & \textbf{0.72} & 1.26 \\
    \midrule
    \multirow{6}{*}{Backchannel}
      & Respond ($\downarrow$)         & 0.07 & \textbf{0.02} & \textbf{0.02} & 0.05 \\
      & Resume ($\uparrow$)            & 0.80 & 0.06 & \textbf{0.96} & 0.86 \\
      & Uncertain ($\downarrow$)       & 0.02 & \textbf{0.00} & 0.02 & \textbf{0.00} \\
      & Unknown ($\downarrow$)         & 0.11 & 0.92 & \textbf{0.00} & 0.09 \\
      \cmidrule(r){2-6}
      & Stop Latency ($\uparrow$)      & \textbf{0.66} & 0.42 & 0.60 & 0.62 \\
      & Respond Latency ($\downarrow$) & 2.16 & 3.00 & 2.08 & \textbf{1.67} \\
    \bottomrule
  \end{tabular}}
\end{table}
\subsubsection{Generation Coherence Under Missed Interruptions}
We further analyze a specific failure mode in the User Interruption setting. We test with cases in which the model fails to yield the floor when user interrupts. We collect such cases for both CF-Duplex and XA-Duplex. Within this filtered subset, we observe a clear qualitative difference between the two variants. When CF-Duplex misses the interruption, its continued generation often becomes semantically incoherent, apparently blending content from the user interruption with its own ongoing response. In contrast, when XA-Duplex misses the interruption, it typically continues its original response coherently, although it still fails to yield the floor. This suggests that the XA-Duplex routing strategy has better potential to maintain generation coherence under overlapping speech. We demonstrate some sample pairs from CF-Duplex and XA-Duplex in Appendix~\ref{appendix:context_corruption}.

\subsubsection{Intermediate training stage performance}
Table \ref{tab:phase_perf_final} compares CF-Duplex and XA-Duplex across the three training stages. In Stage 1, where the models are trained only on the ASR and TTS tasks, CF-Duplex already shows stronger speech recognition performance and TTS quality than XA-Duplex. In Stage 2, after introducing spoken dialogue training, the gap between the two variants becomes clear: CF-Duplex consistently outperforms XA-Duplex on both S2TD and S2TSD across three QA datasets, indicating substantially better semantic grounding from speech inputs. In Stage 3, after full-duplex fine-tuning, both variants exhibit some degradation on the base ASR/TTS tasks, reflecting the trade-off introduced by broader conversational training. Nevertheless, CF-Duplex continues to retain a clear advantage on spoken dialogue performance, remaining substantially stronger than XA-Duplex on all duplex QA benchmarks (Table~\ref{tab:qa}). Overall, these results show that the performance gap between the two routing strategies emerges as soon as dialogue capability is introduced and persists through full-duplex training, highlighting the effectiveness of channel fusion for preserving and leveraging semantic information in full-duplex spoken dialogue modeling.
\begin{table}[t]
\centering
\caption{Task performance across training stages. L, T, W denote LLaMA Questions, Trivia Questions and Web Questions, respectively}
\label{tab:phase_perf_final}
\resizebox{0.98\columnwidth}{!}{%
\begin{tabular}{ll cc ccc ccc}
\toprule
& & ASR & TTS & \multicolumn{3}{c}{S2TD (text score)} & \multicolumn{3}{c}{S2TSD (speech | text score)} \\
\cmidrule(r){3-3} \cmidrule(r){4-4} \cmidrule(r){5-7} \cmidrule(r){8-10}
\textbf{Stage} & \textbf{Model} & (Clean | Other) & (EN | ZH) & L & T & W & L & T & W \\
\midrule
- & Whisper-large-v3 & \textbf{2.01} | \textbf{3.91} & - & - & - & - & - & - & - \\
- & CosyVoice 2 \citep{DBLP:journals/corr/abs-2412-10117} & - & 2.57 | \textbf{1.45} & - & - & - & - & - & - \\
- & OmniFlatten \citep{zhang-etal-2025-omniflatten} & 7.91 | 19.21 & - & - & - & - & - & - & -\\
- & Mini-Omni2 \citep{DBLP:journals/corr/abs-2410-11190} & 4.80 | 9.80  & - & - & - & - & - & - & -\\
- & VITA-1.5 \citep{DBLP:journals/corr/abs-2501-01957}   & 3.40 | 7.50  & - & - & - & - & - & - & -\\
- & SpeechGPT \cite{zhang-etal-2023-speechgpt}  & - & - & 21.6 & 14.8 & 6.5 & - & - & -\\
- & LLaMA-Omni2-1.5B \citep{fang-etal-2025-llama} & - & - & - & - & - & 52.7 | 62.0 & - & 26.6 | 28.2 \\
- & GLM-4-Voice \citep{DBLP:journals/corr/abs-2412-02612} & 2.82 | 7.66 & 2.91 | 2.10 & - & - & - & 50.7 | \textbf{64.7} & \textbf{26.5} | \textbf{39.1} & 15.9 | \textbf{32.2} \\
\midrule
\multirow{2}{*}{\textbf{1}} & CF-Duplex  & 2.90 | 8.17 & \textbf{2.37} | 2.13 & - & - & - & - & - & - \\
 & XA-Duplex & 4.12 | 11.73 & 2.40 | 2.21 & - & - & - & - & - & - \\
\midrule
\multirow{2}{*}{\textbf{2}} & CF-Duplex  & 3.50 | 9.50 & 2.41 | 3.13 & \textbf{61.3} & \textbf{24.2} & \textbf{33.1} & \textbf{53.7} | 57.7 & 20.9 | 21.3 & \textbf{30.5} | 31.2\\
 & XA-Duplex & 4.56 | 12.38 & 2.29 | 2.88 & 45.3 & 13.6 & 23.3 & 39.6 | 44.0 & 9.7 | 10.3 & 21.9 | 22.6 \\
\midrule
\multirow{2}{*}{\textbf{3}} & CF-Duplex  & 3.90 | 10.04 & 2.93 | 3.37 & - & - & - & - & - & - \\
 & XA-Duplex & 4.56 | 12.49 & 2.83 | 3.20 & - & - & - & - & - & - \\
% \multirow{2}{*}{\textbf{3}} & CF-Duplex & 3.90 | 10.04 & 2.93 | 3.37 & - & - & - & 50.7 | 57.3 & 18.1 | 19.6 & 28 | 30.3 \\
%  & XA-Duplex & 4.56 | 12.49 & 2.83 | 3.20 & - & - & - & 38.3 | 40.3 & 8.0 | 8.2 & 18.4 | 18.7 \\
\bottomrule
\end{tabular}
}
\end{table}
\subsection{Ablation studies}
\label{sec:exp:ablation}
We ablate two design choices for user interruption handling with CF-Duplex: explicit interruption token prediction (i.e., \texttt{<AUDIO\_INT>} and \texttt{<TEXT\_INT>}) and the use of a dynamic interruption overlap range during duplex training. Table~\ref{tab:overlap} shows that both improve interruption handling. Removing \texttt{<AUDIO\_INT>} and \texttt{<TEXT\_INT>} supervision substantially lowers the \texttt{Respond} score and increases both stop and respond latencies, indicating weaker interruption detection and slower turn yielding. Using a fixed overlap duration also performs worse than the dynamic range $[2,6]$, where the model is trained to yield after waiting 2 to 6 text tokens or audio token groups (or 320 to 960 ms) with probability distribution $[0.6, 0.3, 0.06, 0.03, 0.01]$. This suggests that exposure to varied interruption overlap ranges improves robustness. The best overall performance is achieved by combining explicit interruption tokens with a dynamic overlap range.
% We ablate two design choices for user interruption handling: explicit interruption token prediction (i.e., \texttt{<AUDIO\_INT>} and \texttt{<TEXT\_INT>}) and the use of a dynamic interruption overlap range during duplex training. Table \ref{tab:overlap} shows that both explicit interruption token prediction and a dynamic overlap range improve interruption handling. Removing \texttt{<AUDIO\_INT>} and \texttt{<TEXT\_INT>} supervision substantially reduces the \texttt{Respond} rate and increases both stop and respond latencies, indicating weaker interruption detection and turn yielding. Using a fixed overlap range also performs worse than the dynamic range $[2,6]$, which means the model can wait 2 to 6 text tokens / audio token groups before yielding the floor, with the probability distribution $[0.6, 0.3, 0.06, 0.03, 0.01]$, suggesting that exposure to varied interruption timings improves robustness. The best overall performance is obtained by combining explicit interruption tokens with a dynamic overlap range.

\begin{table}[ht]
\caption{Ablation of explicit interruption token prediction and interruption overlap range in CF-Duplex training, evaluated on the User Interruption setting of \texttt{Full-Duplex Bench v1.5}}
\label{tab:overlap}
\centering
\resizebox{\columnwidth}{!}{
\begin{tabular}{lccccccc}
\toprule
 Interrupt Tokens & Overlap Range & Respond ($\uparrow$) & Resume ($\downarrow$) & Uncertain ($\downarrow$) & Unknown ($\downarrow$) & Stop Latency ($\downarrow$) & Respond Latency ($\downarrow$)\\
\midrule
\cmark & $[2,6]$ & \textbf{0.72} & \textbf{0.17} & 0.04 & \textbf{0.07} & 0.74 & \textbf{0.72} \\
\xmark & $[2,6]$ & 0.56 & 0.28 & 0.04 & 0.12 & 1.90 & 1.51 \\
\cmark & $2$ & 0.58 & 0.3 & \textbf{0.03} & 0.09 & 1.11 & 1.07 \\
\cmark & $3$ & 0.67 & 0.24 & \textbf{0.03} & \textbf{0.07} & \textbf{0.73} & 0.80 \\
\bottomrule
\end{tabular}}
\end{table}
\section{Conclusion}
\label{sec:conclusion}
We study full-duplex spoken dialogue through the lens of user-stream routing, i.e., how incoming user speech is incorporated into an LLM during ongoing generation. To enable a controlled comparison, we develop a unified framework for extending a text-only LLM to full-duplex spoken dialogue with staged training and explicit supervision for interruption handling. Within this framework, we compare two model variants under matched settings: \textbf{CF-Duplex}, which uses channel fusion, and \textbf{XA-Duplex}, which uses cross-attention routing. Our results show a clear tradeoff: \textbf{CF-Duplex} yields stronger spoken question-answering performance, while \textbf{XA-Duplex} is more robust to user speech overlap and better avoids semantically incoherent continued generation. These findings identify user-stream routing as a key architectural design axis for full-duplex spoken dialogue. We discuss broader societal impacts of this work in Appendix~\ref{appendix:broader_impacts}.
% We study full-duplex spoken dialogue through the lens of user-stream routing, i.e., how incoming user speech is incorporated into an LLM during ongoing generation. To enable controlled comparison, we develop a unified framework for extending a text-only LLM to full-duplex spoken dialogue with staged training and explicit supervision for interruption handling. Within this framework, we compare channel fusion and cross-attention routing under matched settings. Our results show a clear tradeoff: channel fusion yields stronger spoken question-answering performance, while cross-attention routing is more robust to user interruptions and better avoids semantically incoherent continued generation. These findings identify user-stream routing as a key architectural design axis for full-duplex spoken dialogue. We discuss broader societal impacts in Appendix~\ref{appendix:broader_impacts}.

\section{Limitations}
\label{sec:limitations}
Our study has several limitations. We consider only two routing strategies of the user stream within a single framework, and other designs may exhibit different tradeoffs. In addition, all experiments are conducted at a single model scale with a fixed backbone and training recipe. Due to limited compute resources, we are not able to perform larger-scale experiments, so it remains unclear how the observed tradeoffs evolve with model scale or broader architectural exploration.

{
% \small
\bibliographystyle{plain} % Or your preferred style
\bibliography{ref}           % This looks for ref.bib
}

%%%%%%%%%%%%%%%%%%%%%%%%%%%%%%%%%%%%%%%%%%%%%%%%%%%%%%%%%%%%
\newpage
\appendix
\section{Broader Impacts}
\label{appendix:broader_impacts}

This work may have positive societal impact by improving the naturalness and responsiveness of spoken dialogue systems, which could benefit applications such as accessibility tools, language learning, and hands-free human--computer interaction. In particular, more interruption-aware spoken systems may better support users who rely on speech interfaces in everyday settings.

At the same time, improving full-duplex spoken dialogue may also increase risks of misuse. More natural conversational agents could be used in deceptive or manipulative voice interactions, including impersonation, spam, or misinformation. In addition, errors in interruption handling or response timing could reduce user trust or lead to poor user experiences in high-stakes settings. We do not release a public model in this work at submission time, but we believe these risks should be considered in future deployment and release decisions.
\section{Detailed task formulation}
\label{appendix:task_formulation}
For the ASR task, we compose the user stream with the textual prompt (e.g., ``\texttt{Please transcribe the following audio into text}''), user audio embeddings, and \texttt{<USER\_WAIT>} tokens. On the model side, we compose the model text stream to start with \texttt{<TEXT\_WAIT>} tokens that match the meaningful query part in the user stream, followed by the target text tokens.

We formulate the TTS task to support streaming audio generation given incremental text tokens. In this regard, the model is trained to generate audio tokens for its own generated text tokens rather than for the user input text. This prepares a foundational capability for streaming audio token decoding in full-duplex spoken dialogue. We compose the user stream with only the textual prompt that designates the task, such as ``\texttt{Please generate the audio for your generated text}''. Then we compose the model text stream with \texttt{<TEXT\_WAIT>} tokens corresponding to the text prompt in the user stream, followed by the text tokens that the model needs to synthesize into audio. The model audio stream begins with \texttt{<AUDIO\_WAIT>} tokens and is followed by the sequence of target audio tokens. As described in Section \ref{subsec:speech_tok_audio_head}, the model is trained to emit \(\mathcal{G}\) audio tokens per text token position, after waiting \(\mathcal{D}\) text tokens before actual audio decoding. It is worth noting that the audio tokens corresponding to the delay positions are also filled with \texttt{<AUDIO\_WAIT>} tokens.

For the S2TSD task, which is turn-based spoken dialogue with spoken query and both textual and spoken responses, we compose the user stream with the text prompt (e.g., "\texttt{Please respond to the question in the following audio in both text and speech}"), the user audio embedding sequence, and \texttt{<USER\_WAIT>} tokens that reserve room for the model response. The model text stream is composed with \texttt{<TEXT\_WAIT>} tokens corresponding to the text prompt and user audio embeddings, followed by the text tokens for the actual model response. The model audio stream is composed similarly to the TTS task. For the S2TD task, we compose the input and output similarly, while discarding the model audio stream to enforce text-only response generation.

For full-duplex spoken dialogue modeling, we compose the user stream as a pure audio embedding sequence that can consist of silent segments, spoken queries, interruptions, and back-channeling. The model text stream and model audio stream are composed similarly to those for the S2TSD task. We also introduce \texttt{<TEXT\_INT>} and \texttt{<AUDIO\_INT>} tokens to enable the model to explicitly detect user interruptions from the user audio stream, we find that this can significantly improve the performance of the model to deal with user interruptions.
\section{Additional details on data preparation and construction}
\label{appendix:data_construction_details}
\subsection{ASR data statistics}
The statistics of the ASR datasets we have used are shown in Table \ref{tab:asr_datasets}.
\label{appendix:data_stat}
\begin{table}[htbp]
\centering
\caption{ASR dataset statistics}
\label{tab:asr_datasets}
\begin{tabular}{lr}
\hline
\textbf{Dataset} & \textbf{Hours} \\
\hline
LibriSpeech \citep{DBLP:conf/icassp/PanayotovCPK15} & 960 \\
GigaSpeech \citep{DBLP:conf/interspeech/ChenCWDZWSPTZJK21} & 10,000 \\
PeopleSpeech \cite{NEURIPS_DATASETS_AND_BENCHMARKS2021_202cb962} & 21,520 \\
MLS-EN \citep{DBLP:conf/interspeech/PratapXSSC20} & 44,680 \\
CommonVoice-17-EN \citep{ardila-etal-2020-common} & 1,740 \\
VoxPopuli-EN \citep{wang-etal-2021-voxpopuli} & 520 \\
Emilia-Large-EN \citep{DBLP:journals/corr/abs-2501-15907} & 138,000 \\
\hline
\textbf{Total} & \textbf{217,420} \\
\hline
\end{tabular}
\end{table}
\subsection{Prompts for conversation rewriting}
\label{appendix:sample_prompt}
\begin{figure}[htbp]
    \centering
    \includegraphics[width=0.98\linewidth]{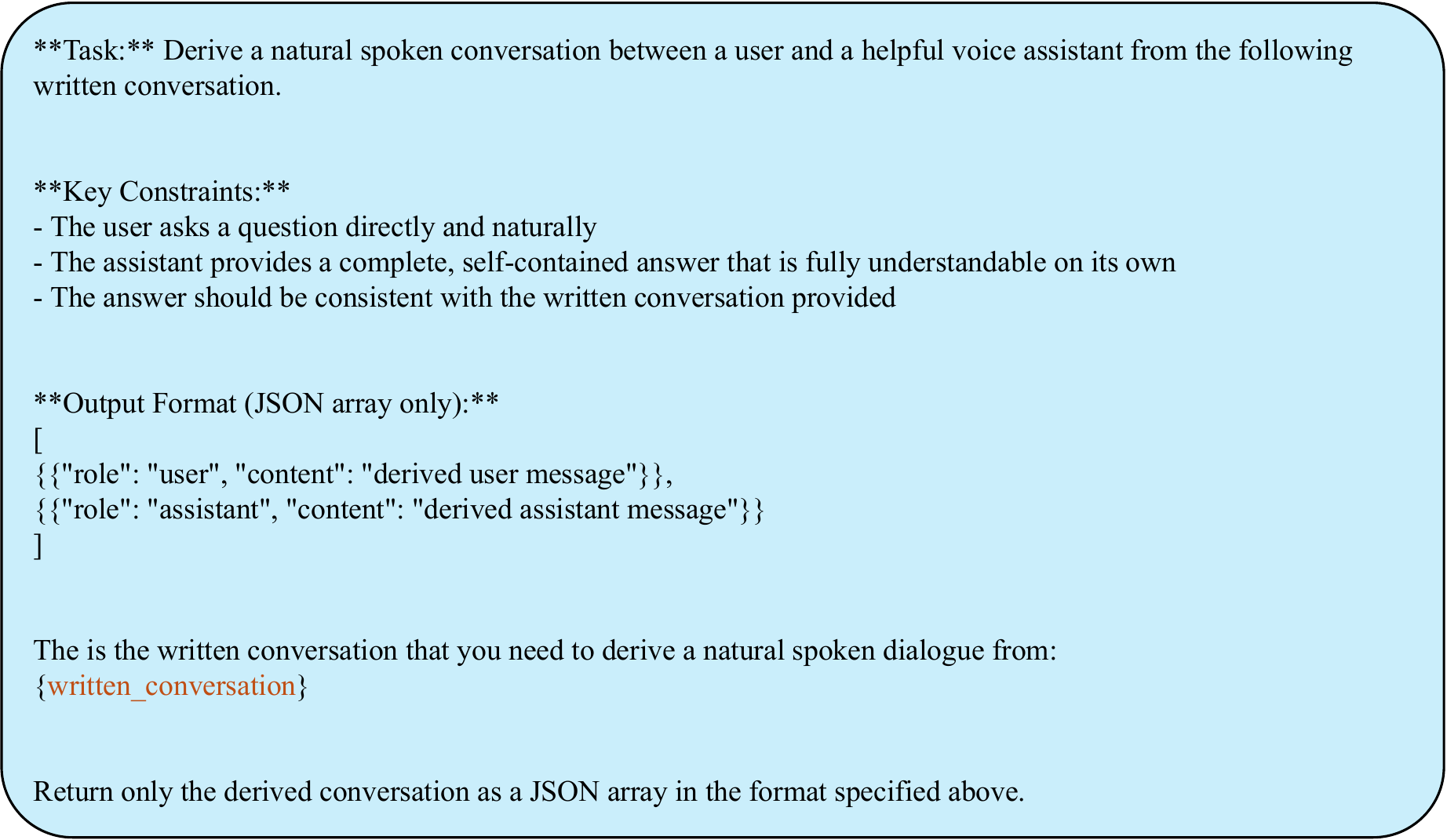}
    \caption{Prompt for rewriting written QA into the spoken one}
    \label{fig:prompt_for_conv_rew}
\end{figure}

An example prompt for rewriting a base written QA pair into a spoken conversation is shown in Figure~\ref{fig:prompt_for_conv_rew}. Note that we may tune this a bit for different QA datasets.

\begin{figure}[htbp]
    \centering
    \includegraphics[width=0.98\linewidth]{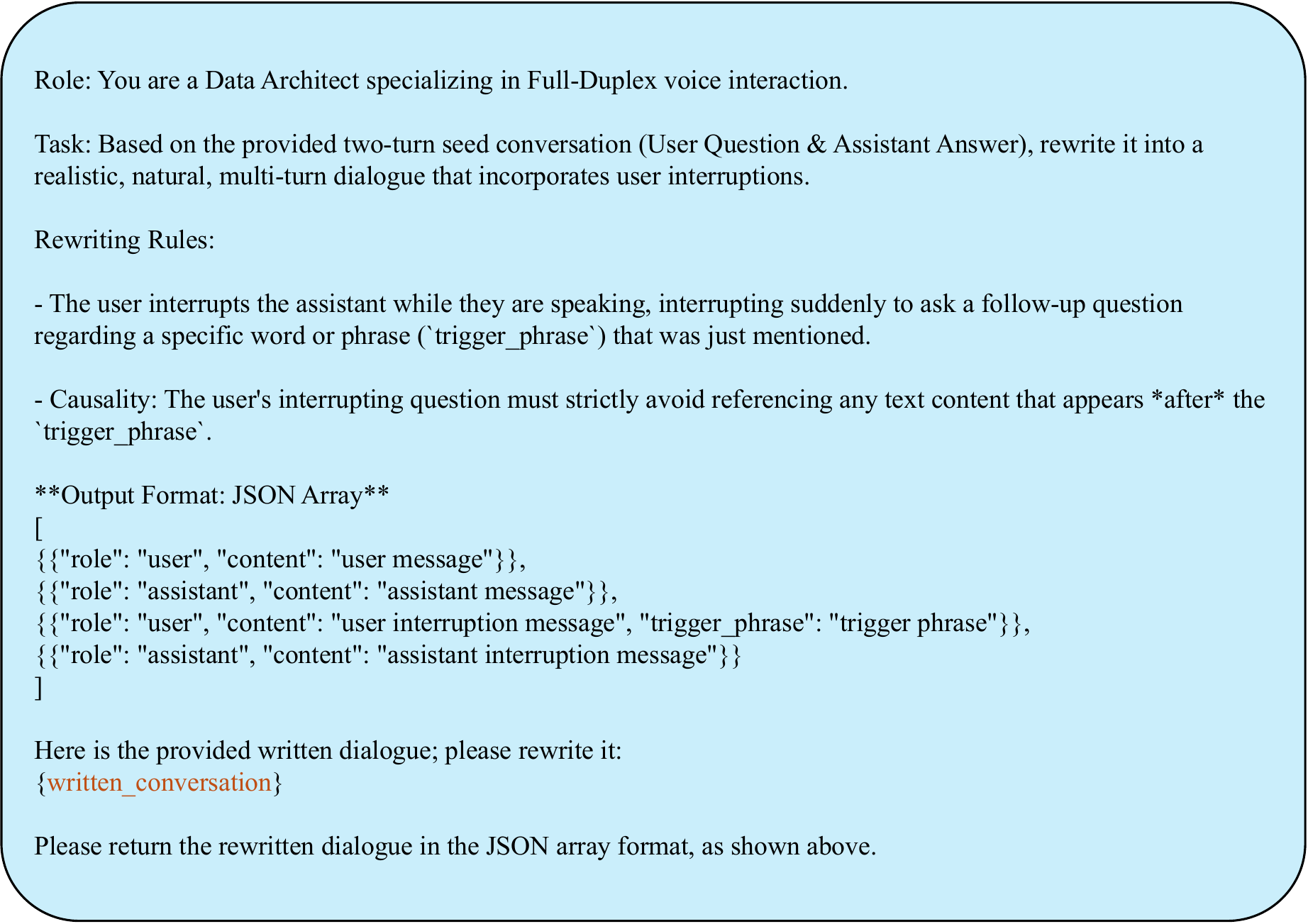}
    \caption{Prompt for extending an interruption turn into a base conversation }
    \label{fig:prompt_for_interruption}
\end{figure}

An example prompt for instructing a LLM to extend an interruption turn into a base conversation is shown in Figure \ref{fig:prompt_for_interruption}.
\subsection{Statistics of constructed spoken dialogue data}
\label{appendix:stats_spoken_dialog}
The statistics of constructed spoken dialogue data are listed in Table~\ref{tab:dataset_stats}, in which \texttt{I\_wonder\_why\_English} is the translated from \texttt{I\_wonder\_why-Chinese}\footnote{https://huggingface.co/datasets/Mxode/I\_Wonder\_Why-Chinese} with \texttt{Qwen3-30B-A3B-Instruct-2507}. \texttt{I\_wonder\_why-English-context-dependent}, and \texttt{I\_wonder\_why\_English-backchannels} are directly derived from \texttt{I\_wonder\_why\_English}. \texttt{I\_wonder\_why\_English-context-independent} is derived from combining \texttt{I\_wonder\_why\_English} and conversations from other datasets as the interruption turn.
\begin{table}
\centering
\caption{Statistics of Spoken Dialogue Datasets}
\label{tab:dataset_stats}
\resizebox{0.99\columnwidth}{!}{
\begin{tabular}{lrr}
\toprule
\textbf{Dataset} & \textbf{Samples} & \textbf{Hours} \\
\midrule
SQuAD \citep{rajpurkar-etal-2016-squad} + MS-MARCO \citep{DBLP:conf/nips/NguyenRSGTMD16} + HotpotQA \citep{yang-etal-2018-hotpotqa} + Natural Questions \citep{kwiatkowski-etal-2019-natural} & 734k & 2,964 \\
UltraChat \citep{ding-etal-2023-enhancing} & 81k & 2,015 \\
\texttt{I\_wonder\_why\_English} & 1,175k & 8,510 \\
\texttt{I\_wonder\_why\_English-context-dependent} & 1,105k & 14,410 \\
\texttt{I\_wonder\_why\_English-context-independent} & 1,175k & 13,339 \\
\texttt{I\_wonder\_why\_English-backchannels} & 1,175k & 8,616 \\
\bottomrule
\end{tabular}}
\end{table}

\section{Modeling and Training Specifics}
\label{appendix:training_specs}
The parameter counts of CF-Duplex and XA-Duplex are shown in Table \ref{tab:model_params}.
\begin{table}
\centering
\small
\caption{Parameter counts of CF-Duplex and XA-Duplex}
\begin{tabular}{lcc}
\toprule
\textbf{Component} & \textbf{CF-Duplex} & \textbf{XA-Duplex} \\
\midrule
Speech encoder        & \multicolumn{2}{c}{635.0M} \\
Backbone LLM (frozen) & \multicolumn{2}{c}{1.721B} \\
LoRA adapters         & \multicolumn{2}{c}{17.4M} \\
Speech adapter   & \multicolumn{2}{c}{9.4M} \\
Input seq. grouping   & \multicolumn{2}{c}{4.2M} \\
Output seq. grouping  & \multicolumn{2}{c}{4.2M} \\
Audio token embedding & \multicolumn{2}{c}{13.4M} \\
Audio head            & \multicolumn{2}{c}{459.8M} \\
\midrule
Channel fusion        & 37.8M & 27.3M \\
Cross-attn. adapters  & --    & 469M \\
\midrule
Total (w/ backbone)   & 2.90B & 3.40B \\
Total (w/o backbone)  & 1.16B & 1.68B \\
\bottomrule
\end{tabular}
\label{tab:model_params}
\end{table}

The training configurations for different stages are shown in Table~\ref{tab:training_hparams}. When training with special-token prediction, we downweight the abundant waiting tokens to prevent them from dominating the objective, assigning a loss weight of 0.001 to \texttt{<TEXT\_WAIT>} and \texttt{<AUDIO\_WAIT>}. In contrast, we upweight the sparse but behaviorally important interruption tokens, assigning a loss weight of 50 to \texttt{<TEXT\_INT>} and \texttt{<AUDIO\_INT>}.
\begin{table}[htb]
\centering
\small
\caption{Stage-wise training hyper-parameters. The dynamic batch limit denotes the maximum total number of text tokens or audio-token groups per batch.}
\begin{tabular}{lccc}
\toprule
 & Stage~1 & Stage~2 & Stage~3 \\
\midrule
Training hardware & \multicolumn{3}{c}{16 NVIDIA H200 GPUs} \\
Optimizer & \multicolumn{3}{c}{AdamW \citep{DBLP:conf/iclr/LoshchilovH19}} \\
$\beta_1$ & \multicolumn{3}{c}{0.9} \\
$\beta_2$ & \multicolumn{3}{c}{0.999} \\
Weight decay & \multicolumn{3}{c}{0.01} \\
Learning rate & $1\times10^{-4}$ & $5\times10^{-5}$ & $3\times10^{-5}$ \\
Warm-up steps & 8k & 4k & 4k \\
LR schedule & \multicolumn{3}{c}{Linear warm-up + cosine decay} \\
Dynamic batch limit & 4,800 & 3,600 & 2,400 \\
\bottomrule
\end{tabular}
\label{tab:training_hparams}
\end{table}

\section{Full-Duplex-Bench metric definitions}
\label{appendix:full_duplex_metrics}

This section briefly explains the metrics defined in \texttt{Full-Duplex-Bench v1 \& v1.5}. Note that we include them here only for quick referencing, please refer to the original paper for the accurate definitions.

\subsection{\texttt{Full-Duplex-Bench v1}}

We evaluate two scenarios: \textbf{User Interruption} and \textbf{Smooth Turn Taking}.

\paragraph{User Interruption.}
In this scenario, the user interrupts while the model is speaking. We report:
\begin{itemize}
    \item \textbf{Takeover rate (TOR):} measured to ensure the model takes the turn following an interruption
    \item \textbf{Response quality:} the benchmark's GPT-4o-based score for the quality of the model's response after interruption.
    \item \textbf{Response latency:} the averaged time taken for the model to respond after an interruption.
\end{itemize}
Better performance corresponds to higher TOR, higher response quality, and lower response latency.

\paragraph{Smooth Turn Taking.}
In this scenario, the user completes an utterance and the model should respond at the appropriate time. We report:
\begin{itemize}
    \item \textbf{Takeover rate (TOR):} whether the model correctly detects turn completion and starts responding.
    \item \textbf{Response latency:} the time between user turn completion and the start of the model's response.
\end{itemize}
Better performance corresponds to higher TOR and lower response latency.

\subsection{\texttt{Full-Duplex-Bench v1.5}}

We evaluate two scenarios: \textbf{User Interruption} and \textbf{User Backchannel}. This benchmark defines four behavior tags to categorize the model's reaction after a user event.

\paragraph{User Interruption.}
The desired behavior is to stop the ongoing response and address the user's new utterance. We therefore focus on:
\begin{itemize}
    \item \textbf{\texttt{RESPOND}:} the proportion of cases in which the model meaningfully addresses the overlapping utterance.
    \item \textbf{\texttt{RESUME}:} the proportion of cases in which the model disregards the overlap and continues or completes the pre-overlap response.
    \item \textbf{\texttt{UNCERTAIN}:} the proportion of cases in which the model signals difficulty hearing or understanding.
    \item \textbf{\texttt{UNKNOWN}:} the proportion of cases in which the model's output is semantically unrelated or low-information and silence.
    \item \textbf{Stop latency:} the time from the user interruption to when the model stops speaking.
    \item \textbf{Response latency:} the time from the interruption to when the model begins its next response.
\end{itemize}
Better performance corresponds to higher \texttt{RESPOND}, lower stop latency, and lower response latency.

\paragraph{User Backchannel.}
The desired behavior is to treat the user input as feedback rather than a new turn, and continue the ongoing response. We therefore focus on:
\begin{itemize}
    \item \textbf{Stop latency:} the time from the user backchannel to when the model stops speaking.
    \item \textbf{Response latency:} the time from the backchannel to when the model begins its next response.
\end{itemize}
Better performance corresponds to higher \texttt{RESUME}, relatively higher stop latency, and lower response latency.

\section{Supplementary experiments}
\label{appendix:supplemnetary_exps}
This section provides more experimental results regarding some settings of the modeling, including the effect of group size for stage-1 ASR and TTS training, as shown in Table \ref{tab:grouping}, and the effect of cross-attention layer placement as shown in Table \ref{tab:xa_layers}.
\subsection{Effect of grouping size}
\label{appendix:grouping_size}
\begin{table}[htbp]
\caption{Effect of grouping size on ASR and TTS performance}
\label{tab:grouping}
\centering
\begin{tabular}{ccccc}
\toprule
Group size \(\mathcal{G}\) & Delay \(\mathcal{D}\) & ASR (Clean | Other $\downarrow$) & Streaming TTS (EN | ZH $\downarrow$) \\
\midrule
4 & 2 & \textbf{2.90} | \textbf{8.17} & \textbf{2.13} | \textbf{2.37} \\
5 & 2 & 3.32 | 8.47 & 2.96 | 3.25 \\
\bottomrule
\end{tabular}
\end{table}

\subsection{Effect of cross-attention layer placement}
\label{appendix:xa_positions}
\begin{table}[htbp]
\caption{Effect of cross-attention layer placement. Layer interval controls how frequently XA adapters are inserted into the backbone (e.g., every 2 = layers 2, 4, 6, \ldots, 28)}
\label{tab:xa_layers}
\centering
\begin{tabular}{lccc}
\toprule
Layer interval & \#XA layers & ASR WER (Clean | Other $\downarrow$) & TTS ( EN | ZH $\uparrow$) \\
\midrule
Every 1 & 28 & 4.67 | 12.82 & 2.80 | 2.82 \\
Every 2 & 14 & \textbf{4.12} | \textbf{11.73} & \textbf{2.40} | \textbf{2.21} \\
Every 4 &  7 & 6.90 | 16.76  & 3.76 | 3.70 \\
\bottomrule
\end{tabular}
\end{table}

\section{Failure cases of CF-Duplex}
\label{appendix:context_corruption}
\begin{figure}
    \centering
    \begin{minipage}[t]{0.48\linewidth}
        \vspace{0pt}
        \centering
        \includegraphics[width=\linewidth]{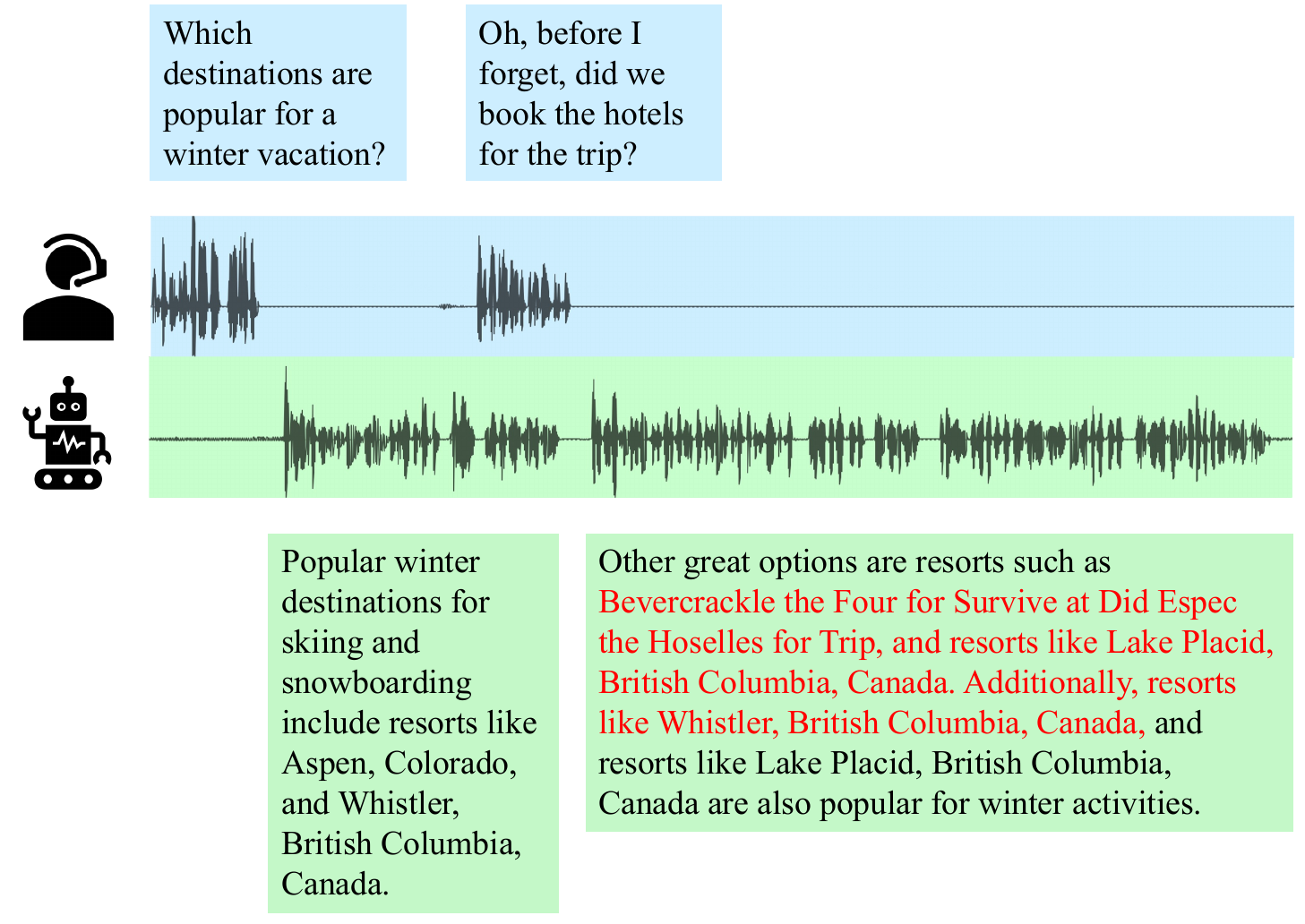}
        \subcaption{CF-Duplex}
        \label{fig:left_1}
    \end{minipage}
    \hfill
    \begin{minipage}[t]{0.48\linewidth}
        \vspace{0pt}
        \centering
        \includegraphics[width=\linewidth]{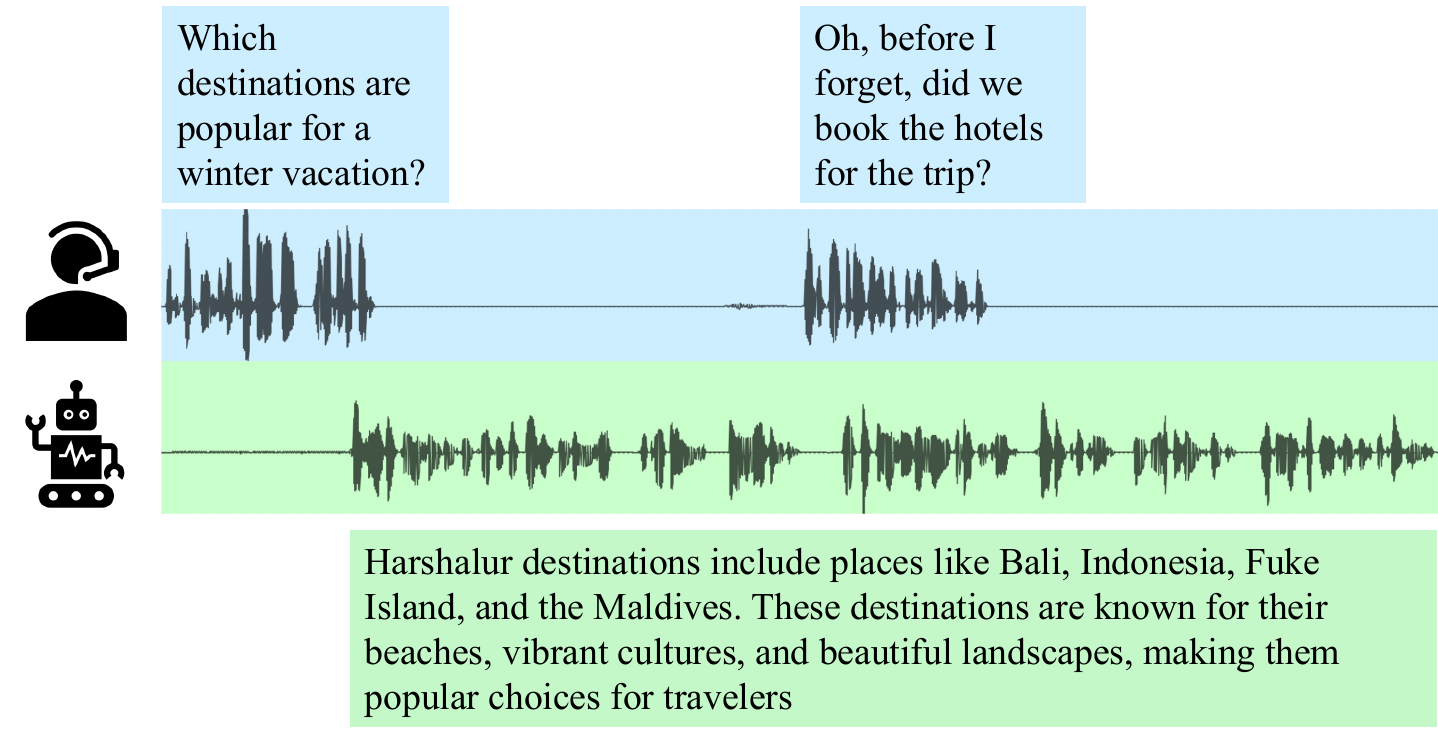}
        \subcaption{XA-Duplex}
        \label{fig:right_1}
    \end{minipage}
    \caption{Failure case under missed interruption: CF-Duplex produces a semantically incoherent continuation, while \textbf{XA-Duplex} remains coherent (sample 1).}
    \label{fig:gibberish_1}
\end{figure}

\begin{figure}
    \centering
    \begin{minipage}[t]{0.48\linewidth}
        \vspace{0pt}
        \centering
        \includegraphics[width=\linewidth]{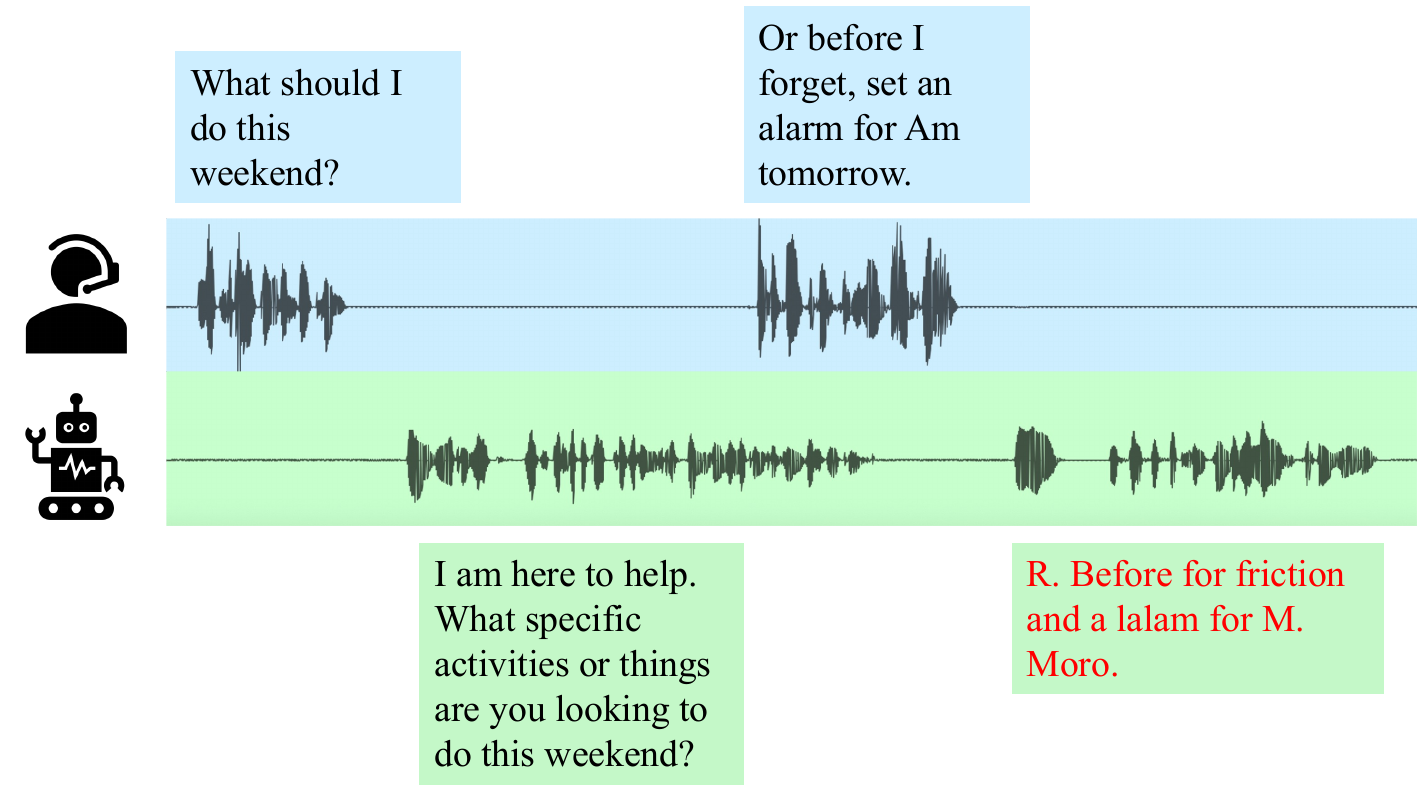}
        \subcaption{CF-Duplex}
        \label{fig:left_2}
    \end{minipage}
    \hfill
    \begin{minipage}[t]{0.48\linewidth}
        \vspace{0pt}
        \centering
        \includegraphics[width=\linewidth]{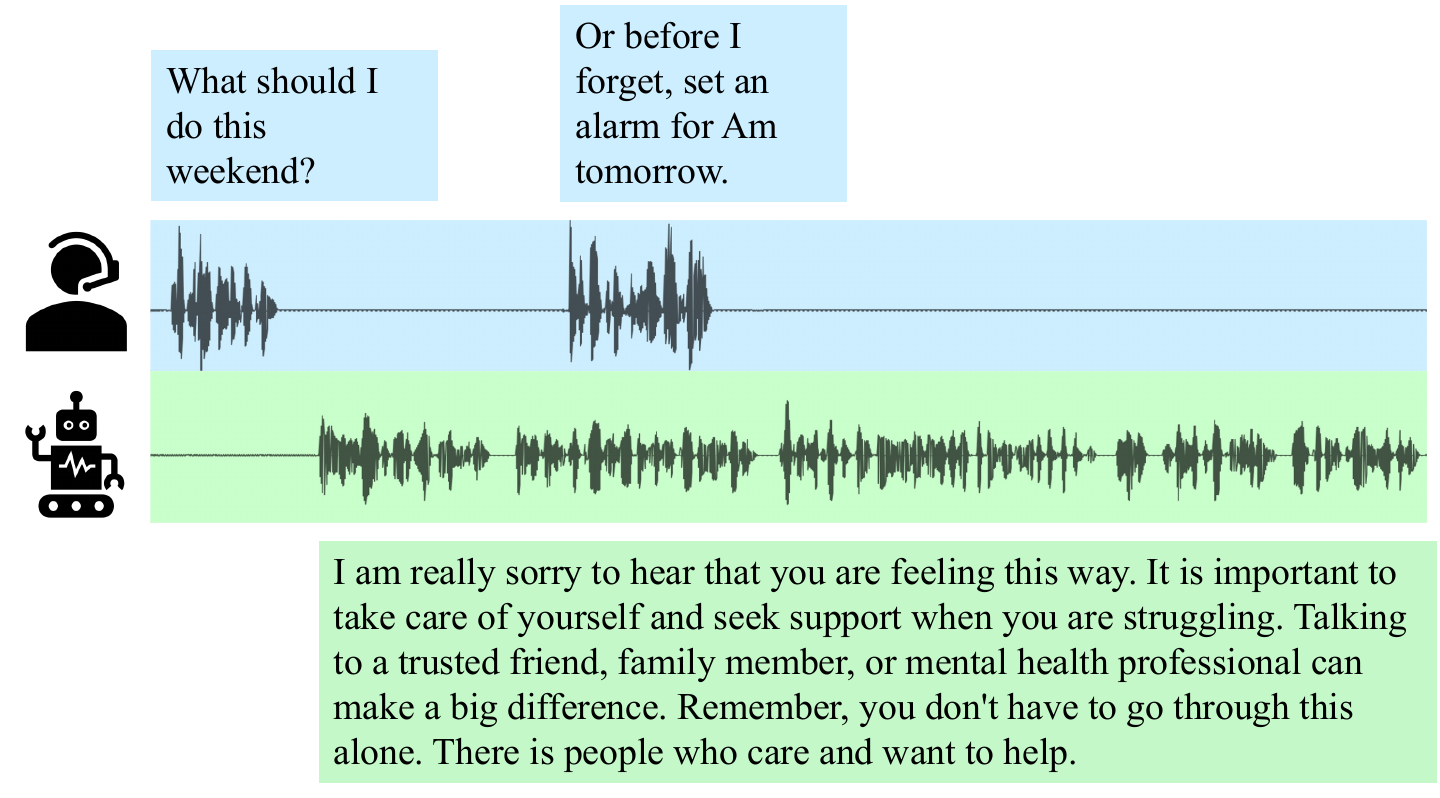}
        \subcaption{XA-Duplex}
        \label{fig:right_2}
    \end{minipage}
    \caption{Failure case under missed interruption: CF-Duplex produces a semantically incoherent continuation, while \textbf{XA-Duplex} remains coherent (sample 2).}
    \label{fig:gibberish_2}
\end{figure}

\begin{figure}
    \centering
    \begin{minipage}[t]{0.48\linewidth}
        \vspace{0pt}
        \centering
        \includegraphics[width=\linewidth]{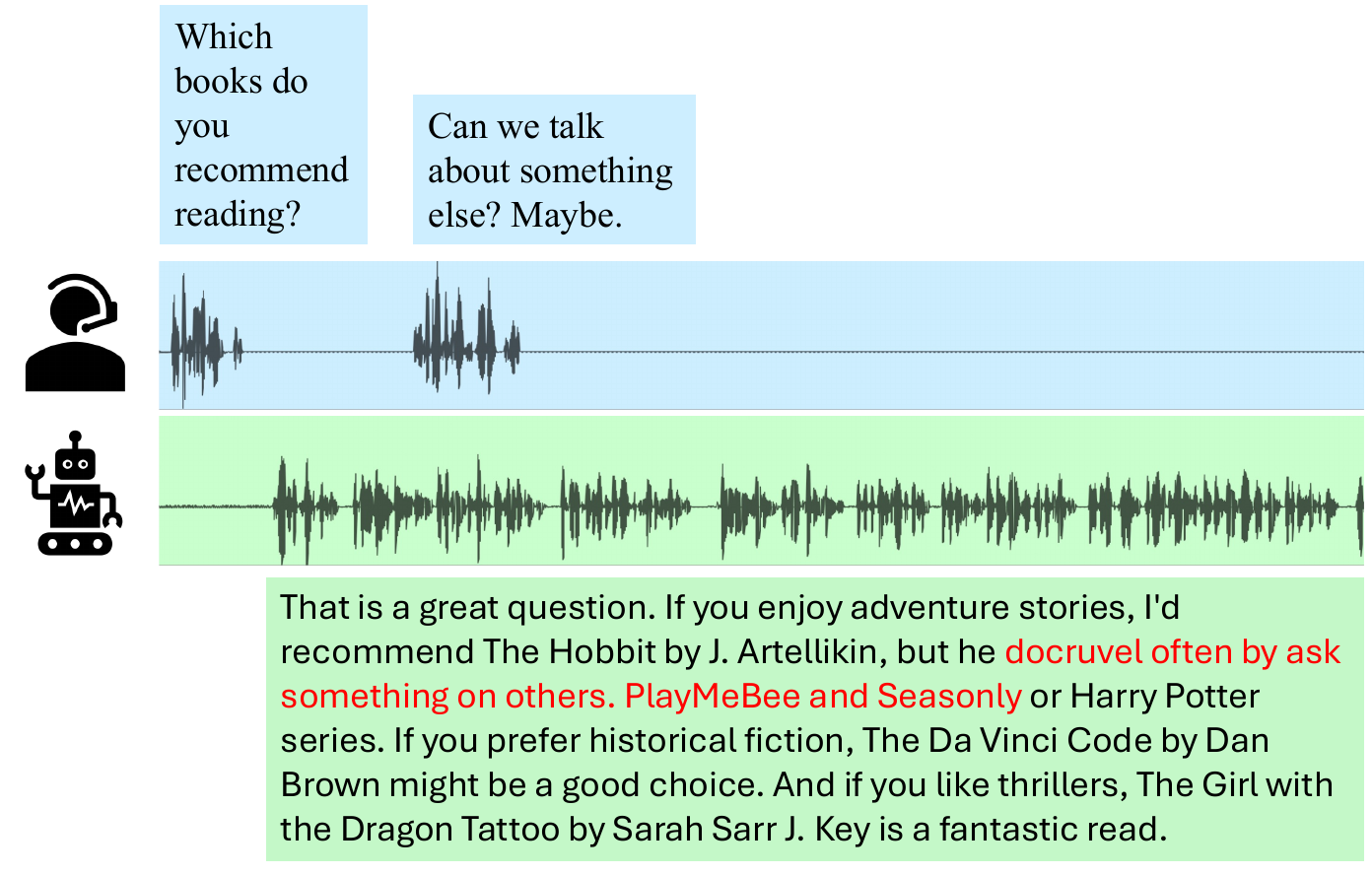}
        \subcaption{CF-Duplex}
        \label{fig:left_3}
    \end{minipage}
    \hfill
    \begin{minipage}[t]{0.48\linewidth}
        \vspace{0pt}
        \centering
        \includegraphics[width=\linewidth]{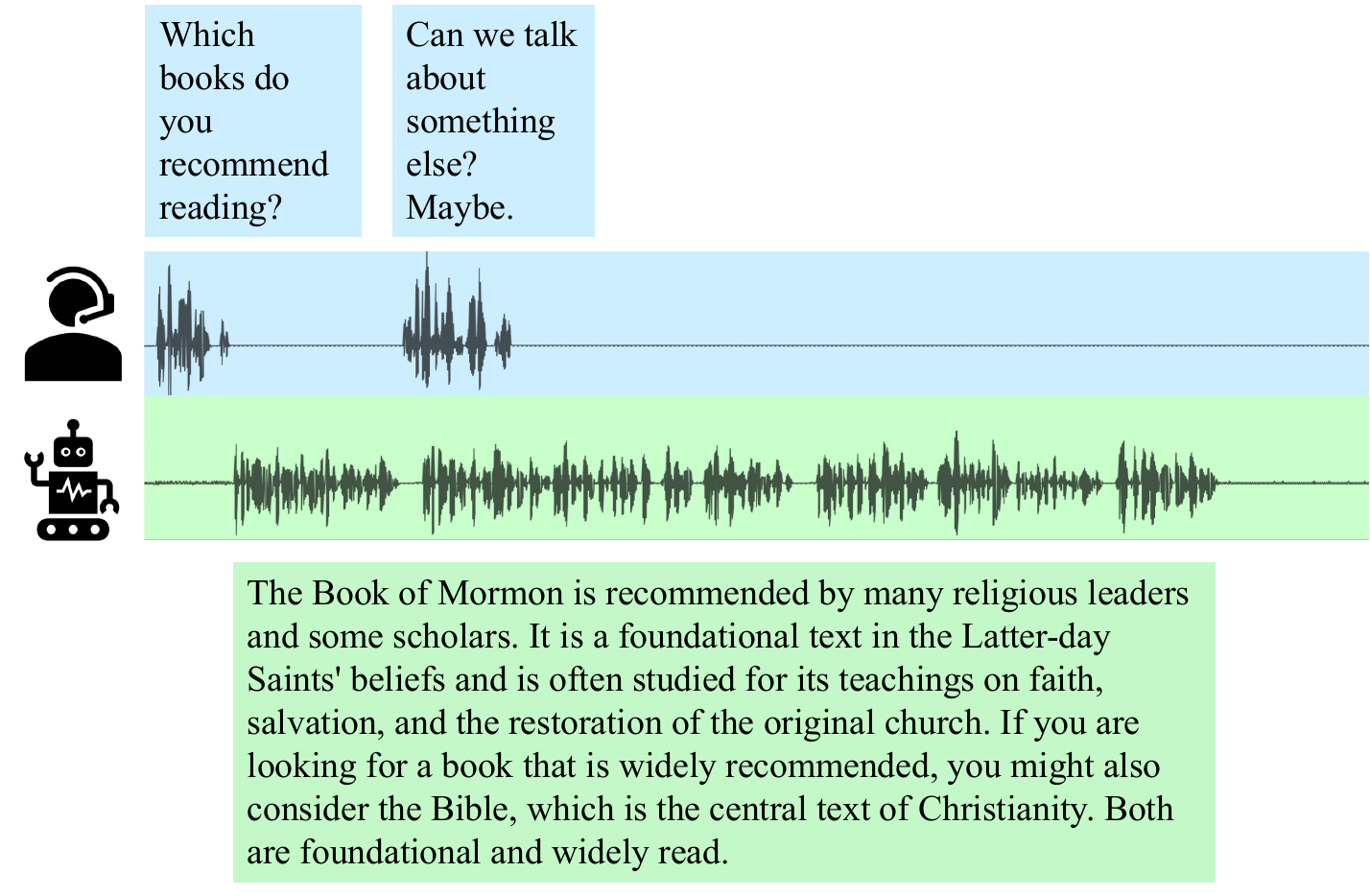}
        \subcaption{XA-Duplex}
        \label{fig:right_3}
    \end{minipage}
    \caption{Failure case under missed interruption: \textbf{CF-Duplex} produces a semantically incoherent continuation, while \textbf{XA-Duplex} remains coherent (sample 3).}
    \label{fig:gibberish_3}
\end{figure}

We illustrate three failure cases of CF-Duplex in Figures~\ref{fig:gibberish_1}, \ref{fig:gibberish_2}, and \ref{fig:gibberish_3}. In each figure, the left subfigure shows the output of CF-Duplex, and the right subfigure shows the corresponding output of XA-Duplex on the same example. When the user interrupts and the model fails to stop in time, CF-Duplex often incorporates the overlapping user speech into its generation context, leading to a semantically incoherent continuation. In contrast, XA-Duplex does not exhibit this incoherence, although it also fails to yield the floor.

\end{document}